\documentclass{article}

\usepackage{microtype}
\usepackage{graphicx}
\usepackage{subfigure}
\usepackage{booktabs} 
\usepackage{multirow}
\usepackage{tablefootnote}
\usepackage{hyperref}

\usepackage{hyperref}

\usepackage[accepted]{icml2024}

\usepackage{amsmath}
\usepackage{amssymb}
\usepackage{mathtools}
\usepackage{amsthm}
\usepackage{graphicx}
\usepackage{dcolumn}
\usepackage[capitalize,noabbrev]{cleveref}

\theoremstyle{plain}

\theoremstyle{definition}

\theoremstyle{remark}

\usepackage[textsize=tiny]{todonotes}

\icmltitlerunning{FreeBind: Free Lunch in Unified Multimodal Space via Knowledge Fusion}

\begin{document}

\twocolumn[
\icmltitle{FreeBind: Free Lunch in Unified Multimodal Space via Knowledge Fusion}


\icmlsetsymbol{equal}{*}

\begin{icmlauthorlist}
\icmlauthor{Zehan Wang}{equal,yyy}
\icmlauthor{Ziang Zhang}{equal,yyy}
\icmlauthor{Xize Cheng}{yyy}
\icmlauthor{Rongjie Huang}{yyy}
\icmlauthor{Luping Liu}{yyy}
\icmlauthor{Zhenhui Ye}{yyy}
\icmlauthor{Haifeng Huang}{yyy}
\icmlauthor{Yang Zhao}{xxx}
\icmlauthor{Tao Jin}{yyy}
\icmlauthor{Peng Gao}{zzz}
\icmlauthor{Zhou Zhao}{yyy}
\end{icmlauthorlist}

\icmlaffiliation{yyy}{Zhejiang University}
\icmlaffiliation{xxx}{ByteDance}
\icmlaffiliation{zzz}{Shanghai AI Lab}

\icmlcorrespondingauthor{Zhou Zhao}{zhaozhou@zju.edu.cn}

\icmlkeywords{Machine Learning, ICML}

\vskip 0.3in
]



\printAffiliationsAndNotice{\icmlEqualContribution} 

\begin{abstract}
Unified multi-model representation spaces are the foundation of multimodal understanding and generation. However, the billions of model parameters and catastrophic forgetting problems make it challenging to further enhance pre-trained unified spaces. In this work, we propose \textbf{FreeBind}, an idea that treats multimodal representation spaces as basic units, and freely augments pre-trained unified space by integrating knowledge from extra expert spaces via ``space bonds". Specifically, we introduce two kinds of basic space bonds: 1) \textbf{Space Displacement Bond} and 2) \textbf{Space Combination Bond}. Based on these basic bonds, we design \textbf{Complex Sequential \& Parallel Bonds} to effectively integrate multiple spaces simultaneously. Benefiting from the modularization concept, we further propose a coarse-to-fine customized inference strategy to flexibly adjust the enhanced unified space for different purposes. Experimentally, we bind ImageBind with extra image-text and audio-text expert spaces, resulting in three main variants: ImageBind++, InternVL$_{I\!B}$ and InternVL$_{I\!B}$++. These resulting spaces outperform ImageBind on 5 audio-image-text downstream tasks across 9 datasets. Moreover, via customized inference, it even surpasses the advanced audio-text and image-text expert spaces. Our code and checkpoints will be released at \url{https://github.com/zehanwang01/FreeBind}
\end{abstract}

\section{Introduction}
Unified multimodal representation aims to learn a semantically shared representation space for many modalities (such as audio, image, language and 3D point cloud)~\cite{girdhar2023ImageBind,wang2023one,guzhov2022audioclip, wu2022wav2clip,wu2023large, xue2023ulip,xue2023ulip2,liu2023openshape}. As an important foundation for multimodal understanding~\cite{liu2023visual,zhu2023minigpt,wang2023chat,han2023ImageBind} and generation~\cite{huang2023make,tang2023any,wu2023next,rombach2022high,podell2023sdxl}, a unified multimodal space is crucial for artificial general intelligence.

Existing advanced unified multimodal representation space~\cite{zhu2023languagebind,girdhar2023ImageBind,zhou2023uni3d} are built on billion-level data and parameters. Learning such a unified space demands exceedingly costly computational resources, and further enhancing the pre-trained space often requires huge training resources or faces the catastrophic forgetting problem. These challenges limit the further development of unified multimodal representation.



\begin{figure}[t]
	\centering
	\includegraphics[width=0.85\linewidth]{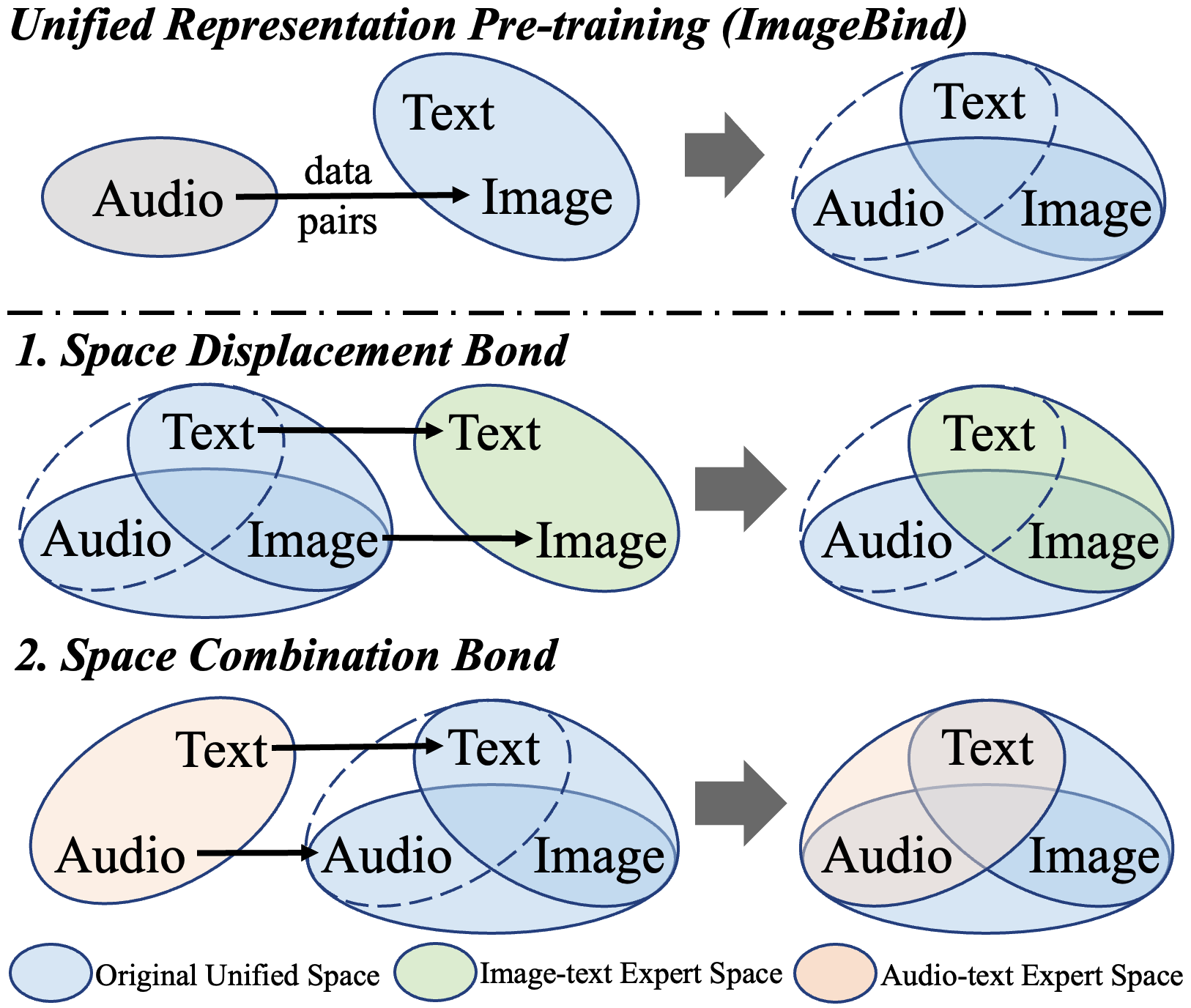}
    \vspace{-0.8\baselineskip}
	\caption{High-level overview of FreeBind. We propose two basic kinds of space bonds: space displacement bond and space combination bond, to efficiently augment unified space by integrating knowledge of extra expert spaces.}
    \vspace{-1\baselineskip}
        \label{fig:overview}
\end{figure}

In this paper, we propose \textbf{FreeBind}, an efficient knowledge fusion scheme to enhance pre-trained unified space. Specifically, we propose to bind unified space (i.e., space for many modalities) with expert space (i.e., space focus on single modalities pair) via two basic ``space bonds":

1) \textbf{Space Displacement Bond.} We align the unified space to the expert space to inherit all the knowledge of the expert space. However, remapping the entire unified space compromises the knowledge of unified space. Additionally, when integrating multiple expert spaces, cascaded displacements are susceptible to cumulative errors. Overall, displacement bond is a radical knowledge fusion solution that sacrifices some information from the unified space in exchange for full expert knowledge.

2) \textbf{Space Combination Bond.} Complementary to displacement bond, we also propose a moderate knowledge fusion scheme called combination bond, which aligns expert space to unified space. Since unified space is frozen, its knowledge can be preserved and we can combine multiple expert spaces in parallel. However, as the expert space is reprojected, the combination bond can only partially integrate the knowledge of expert space.

Based on these two complementary basic bonds, we further propose \textbf{Complex Sequential \& Parallel bonds} to effectively integrate multiple expert spaces simultaneously. Specifically, due to the pivotal role of image-text representations in unified spaces, we first integrate the unified space with advanced image-text expert space via displacement bond and tune the product to repair its lost knowledge. Then, we combine extra expert spaces via combination bond in parallel to further enhance the unified space. For the final resulting space, we design a coarse-to-fine customized inference strategy to flexibly suit different applications by selecting modules and adjusting combining factors.


To demonstrate the effectiveness of FreeBind, we study practical application on the audio-image-text unified space of ImageBind~\cite{han2023ImageBind}. By integrating one image-text and two audio-text expert spaces, we construct state-of-the-art audio-image-text space that significantly surpasses ImageBind. Furthermore, leveraging the flexibility of customized inference, we achieve even better performance in image-text or audio-text tasks than the source expert spaces.



Our contributions can be summarized as follows:
\begin{itemize}
    \item We present \textbf{FreeBind}, an approach that conceptualizes multimodal spaces as basic unit and fuses the knowledge of multimodal representation spaces through space bonds.
    \item We propose two complementary basic bonds between two spaces: displacement and combination bond. Building on these foundations, we further introduce complex sequential \& parallel bonds for integrating multiple spaces simultaneously.
    \item We design a simple yet effective projector learning pipeline and propose a mixture-of-projectors strategy to strengthen the robustness of space alignments.
    \item We employ \textbf{FreeBind} on ImageBind to verify its effectiveness. By integrating advanced image-text and audio-text expert spaces, we establish a state-of-the-art audio-image-text space with limited resources.
\end{itemize}

\section{Related work}
\subsection{Multimodal Representation Space}
Multimodal representation space aims to embed different modality inputs into a joint space. Recent multimodal space research mainly focuses on two aspects: building stronger alignment between two modalities (i.e., expert spaces) or enabling more modalities input (i.e., unified spaces).

Current expert space achieves impressive performance on various modality pairs. By collecting a large collection of image-text pairs, CLIP~\cite{radford2021learning} and ALIGN~\cite{jia2021scaling} show impressive performance and generalization ability. The recent InternVL~\cite{chen2023internvl} scale up the visual encoder to 6 billion parameters and achieves the most advanced performance on most vision-language downstream tasks. The success of vision-language representation inspires more research to explore contrastive representation on other modality pairs. CLAP~\cite{wu2023large} learns high-quality audio-text representation space via massive audio-text pairs, while VideoCLIP~\cite{xu2021videoclip} obtains shared video and text representations from video-text data. In addition to general multimodal representations, some recent researches~\cite{zhang2022contrastive,CLAP2022,CLAP2023} attempt to develop domain-specific pre-trained multimodal spaces, such as music or speech versions of CLAP~\cite{wu2023large}, and image-text space specifically learned on medical images~\cite{zhang2022contrastive}. 

On the other hand, many recent works have tried to develop a unified representation space for more than three modalities to support more diverse applications. These unified space learning approaches collect massive multimodal data pairs and train encoders to align new modalities with a pre-trained image-text space. AudioCLIP~\cite{guzhov2022audioclip} and WAV2CLIP~\cite{wu2022wav2clip} align audio inputs to CLIP by constructing audio-text-image data. Recent ImageBind~\cite{girdhar2023ImageBind} collects and organizes image-paired data of four modalities, and learns encoders of these modalities that aligned to CLIP space. Similarly, LanguageBind~\cite{zhu2023languagebind} align encoders of different modalities to CLIP via constructing language paired data.

Our method aims to integrate the knowledge of expert spaces into a pre-trained unified space, thereby enhancing the unified space with limited resources and enabling it to benefit from breakthroughs of expert spaces. Moreover, via customizing the inference process, the augmented unified space can even surpass expert spaces in terms of their expertise.

\subsection{Knowledge Fusion in Multimodal Representation}
Recent C-MCR~\cite{wang2023connecting} and Ex-MCR~\cite{wang2023extending} first study how to learn new knowledge by integrating multiple expert spaces. Specifically, C-MCR builds expert space by connecting two expert spaces with one shared modality. Subsequently, Ex-MCR proposes extending one space to another instead of connecting both to build a new one. This extending paradigm facilitates better modality scalability and can build a unified space by extending multiple expert spaces into a based expert space via their shared modalities.

Although these methods also focus on knowledge fusion in multimodal space, our method is fundamentally different from them. C-MCR and Ex-MCR are specifically designed for expert spaces with one and only one shared modality. Such strict usage requirements limit their application. In contrast, our method aims to augment pre-trained unified spaces with expert spaces, which involve multiple shared modalities and more general application scenarios.


\begin{figure*}[t]
	\centering
	\includegraphics[width=0.90\linewidth]{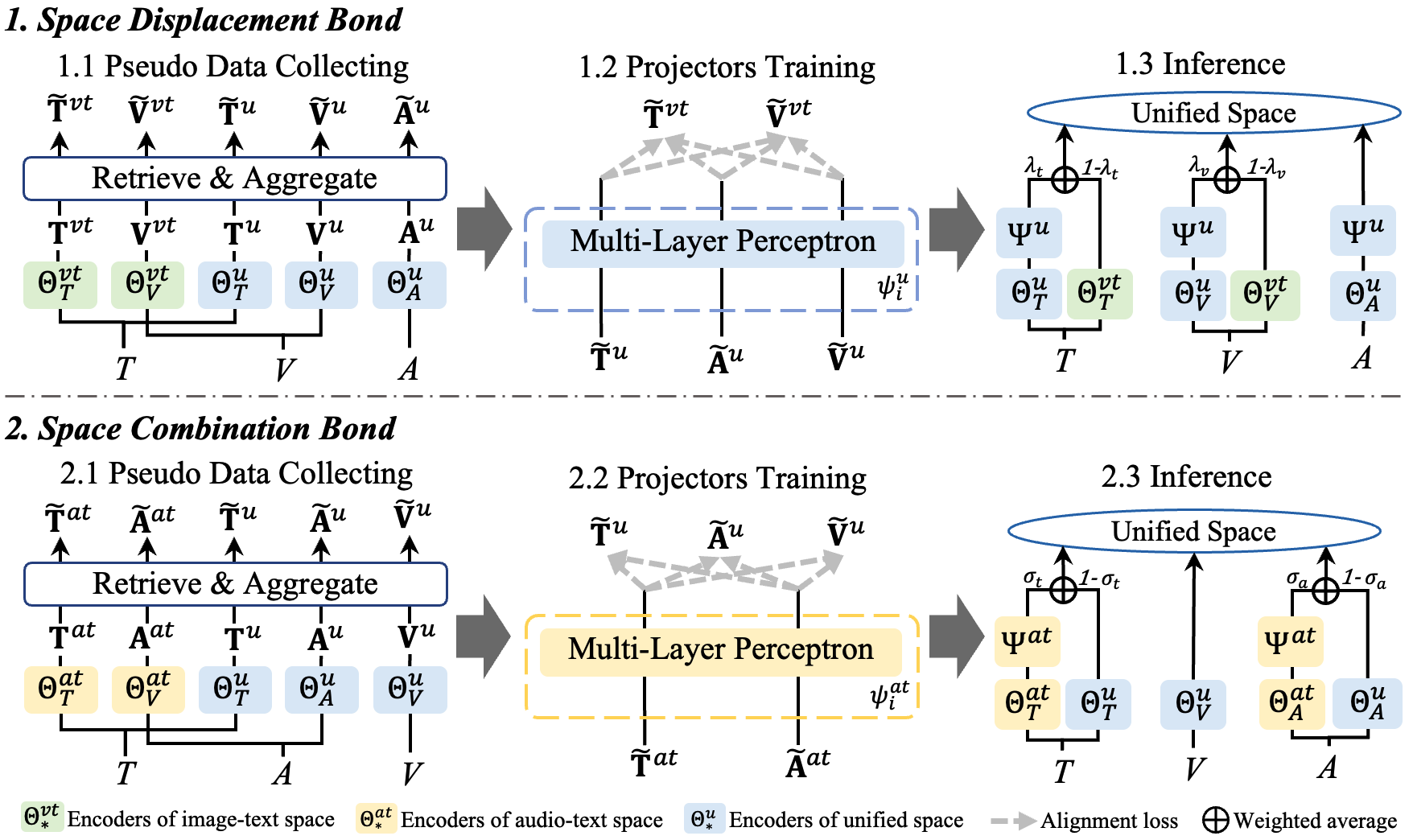}
        \vspace{-0.6\baselineskip}
	\caption{The pipeline of basic space displacement bond and space combination bond.}
        \label{fig:model}
\end{figure*}

\section{Method}
We introduce \textbf{FreeBind}, a training-efficient method designed to enhance pre-trained unified space through knowledge fusion. This section explores its application in augmenting audio-image-text unified space with image-text and audio-text expert spaces. Initially, we formulate the problem, followed by outlining two basic bonds and their composition. Finally, we delve into the customized coarse-to-fine inference strategy.

\subsection{Problem formulation}
The audio-image-text unified space are denoted as $\mathcal{A}^u\mathcal{V}^u\mathcal{T}^u$. Correspondingly, the image-text and audio-text expert spaces can be represented as $\mathcal{V}^{vt}\mathcal{T}^{vt}$ and $\mathcal{A}^{at}\mathcal{T}^{at}$ respectively. The superscripts $^u$, $^{vt}$, and $^{at}$ signify the unified space, image-text and audio-text expert space respectively. With these symbols, the displacement and combination bonds in Figure~\ref{fig:overview} can be expressed as:
\begin{equation}
    \label{Dr}
    \mathcal{A}^u\mathcal{V}^u \mathcal{T}^u\! + \!\operatorname{d}(\mathcal{V}^{vt}\mathcal{T}^{vt}) {\to} \hat{\mathcal{A}}^u(\mathcal{\hat{V}}^u_{1\!-\!\lambda_v}\mathcal{V}^{vt}_{\lambda_v})(\hat{\mathcal{T}}^u_{1\!-\!\lambda_t}\mathcal{T}^{vt}_{\lambda_t})
\end{equation}
\begin{equation}
    \label{Cr}
     \mathcal{A}^u\mathcal{V}^u\mathcal{T}^u\! +\! \operatorname{c}(\mathcal{A}^{at}\mathcal{T}^{at}) {\to} (\mathcal{A}^u_{1\!-\sigma\!_a}\hat{\mathcal{A}}^{at}_{\sigma\!_a})\mathcal{V}^u(\mathcal{T}^u_{1\!-\sigma\!_t}\hat{\mathcal{T}}^{at}_{\sigma\!_t})
\end{equation}
where superscript $\hat{\ }$ means the representations are remapped, $\operatorname{d}({\cdot})$ and $\operatorname{c}({\cdot})$ indicates displacement and combination bond, respectively. (${\lambda_v}, {\lambda_t}$) and (${\sigma\!_a}, {\sigma\!_t}$) are the combining factors of expert spaces. The output spaces in Equation \ref{Dr} and \ref{Cr} are illustrated in the 1.3 and 2.3 part of Figure \ref{fig:model}, and the $(\mathcal{\hat{V}}^u_{1\!-\!\lambda_v}\mathcal{V}^{vt}_{\lambda_v})$ can be formulated as:
\begin{equation}
    (\mathcal{\hat{V}}^u_{1\!-\!\lambda_v}\mathcal{V}^{vt}_{\lambda_v}) = (1\!-\!\lambda_v) \mathcal{\hat{V}}^u + \lambda_v \mathcal{V}^{vt}
\end{equation}
To reflect the pre-trained knowledge of unified space, some unpaired images $V$, texts $T$, and audios $A$ are encoded into the unified space. The corresponding features are denoted as $\mathbf{V}^u \! \in \! \mathbb{R}^{n_v \times d_u}$, $\mathbf{T}^u \! \in \! \mathbb{R}^{n_t \times d_u}$, $\mathbf{A}^u \! \in \! \mathbb{R}^{n_a \times d_u}$, where $d_u$ is the dimension of the unified space. At the same time, the same data is also encoded into expert spaces, serving as bonds between expert spaces and unified space. For image-text expert space, the embeddings are denoted as $\mathbf{V}^{{vt}} \! \in \! \mathbb{R}^{n_v \times d_{vt}}, \mathbf{T}^{vt} \! \in \! \mathbb{R}^{n_t \times d_{vt}}$, while the embeddings in audio-text expert space can be represented as $\mathbf{A}^{at} \! \in \! \mathbb{R}^{n_a \times d_{at}}, \mathbf{T}^{at} \! \in \! \mathbb{R}^{n_t \times d_{at}}$.



\subsection{Basic Space Bonds}

\subsubsection{Pseudo Datasets Collection}
To fuse different multimodal spaces, the initial step involves capturing correlations between different spaces and modalities. To this end, we collect robust and diverse pseudo datasets to bond two different spaces. 

Taking the collection of pseudo datasets collection between image-text expert space and unified space as an example, the embeddings of the expert and unified spaces are $\mathbf{T}^{vt}$, $\mathbf{V}^{vt}$, $\mathbf{T}^u$, $\mathbf{V}^u$ and $\mathbf{A}^u$. The correlation between different modalities can be obtained through the inherent multimodal semantic alignment of embeddings $\mathbf{T}^{vt}$-$\mathbf{V}^{vt}$ and $\mathbf{T}^u$-$\mathbf{V}^u$-$\mathbf{A}^u$ within each space. On the other hand, the correlation between different spaces can be established via the native semantic consistency of $\mathbf{T}^{vt}$-$\mathbf{T}^u$, $\mathbf{V}^{vt}$-$\mathbf{V}^u$ due to the same data source. Combining these two kinds of correlation, we can obtain pseudo multimodal pairs from unpaired or partially-paired data. Furthermore, we retrieve pseudo pairs starting from different modalities respectively, which brings more diverse and comprehensive datasets.

When integrating the unified and image-text spaces, text and image are the shared modalities. The pseudo pairs aggregation process starting from shared modalities (i.e., text and image) can be respectively expressed as:
\begin{equation}
\begin{gathered}
    \label{eq:over-data}
    \tilde{\mathbf{T}}^{vt} = \mathbf{T}^{vt}; \ 
    \tilde{\mathbf{V}}^{vt} = \operatorname{softmax}(\tilde{\mathbf{T}}^{vt} {\mathbf{V}^{vt}}^\top)\mathbf{V}^{vt};
    \\
    \tilde{\mathbf{T}}^u  = \mathbf{T}^u; \ 
    \tilde{\mathbf{V}}^{u} = \operatorname{softmax}(\tilde{\mathbf{T}}^u {\mathbf{V}^{u}}^\top) \mathbf{V}^{u}; 
    \\
    \tilde{\mathbf{A}}^{u} = \operatorname{softmax}(\tilde{\mathbf{V}}^u {\mathbf{A}^{u}}^\top) \mathbf{A}^{u}
\end{gathered}
\end{equation}
\begin{equation}
\begin{gathered}
    \label{eq:over-data}
    \tilde{\mathbf{V}}^{vt} = \mathbf{V}^{vt}; \ 
    \tilde{\mathbf{T}}^{vt} = \operatorname{softmax}(\tilde{\mathbf{V}}^{vt} {\mathbf{T}^{vt}}^\top)\mathbf{T}^{vt};
    \\
    \tilde{\mathbf{V}}^u  = \mathbf{V}^u; \ 
    \tilde{\mathbf{T}}^{u} = \operatorname{softmax}(\tilde{\mathbf{V}}^u {\mathbf{T}^{u}}^\top) \mathbf{T}^{u}; 
    \\
    \tilde{\mathbf{A}}^{u} = \operatorname{softmax}(\tilde{\mathbf{V}}^u {\mathbf{A}^{u}}^\top) \mathbf{A}^{u}
\end{gathered}
\end{equation}
The dataset collection from non-shared modality (i.e., audio) can be formulated as:
\begin{equation}
\begin{gathered}
    \label{eq:over-data}
    \tilde{\mathbf{A}}^u = \mathbf{A}^u; \\
    \tilde{\mathbf{V}}^{u}\! =\! \operatorname{softmax}(\tilde{\mathbf{A}}^u {\mathbf{V}^{u}}^\top\!)\mathbf{V}^{u}; \ 
    \tilde{\mathbf{V}}^{vt}\! =\! \operatorname{softmax}(\tilde{\mathbf{A}}^u {\mathbf{V}^{u}}^\top\!)\mathbf{V}^{vt};
    \\
    \tilde{\mathbf{T}}^{u}\! =\! \operatorname{softmax}(\tilde{\mathbf{V}}^u {\mathbf{T}^{u}}^\top\!) \mathbf{T}^{u}; \ 
    \tilde{\mathbf{T}}^{vt}\! =\! \operatorname{softmax}(\tilde{\mathbf{V}}^{u} {\mathbf{T}^{u}}^\top\!) \mathbf{T}^{vt}
\end{gathered}
\end{equation}
where the superscript $\tilde{\ \ }$ indicates embeddings are processed to be pseudo embedding pairs. The sets of pseudo pairs $(\tilde{\mathbf{T}}^b, \tilde{\mathbf{V}}^b, \tilde{\mathbf{T}}^u, \tilde{\mathbf{V}}^u, \tilde{\mathbf{A}}^u)$ collected from text, image and audio are denoted as $D_T$, $D_V$ and $D_A$, respectively.

When integrating an audio-text expert space with a unified space, the shared modalities are audio and text. The overall pseudo dataset collection process is similar to the above, and the detailed equations can be found in the Appendix.

\subsubsection{Space Alignments}
\paragraph{Single Projector Training}
The previous space alignment methods, C-MCR and Ex-MCR, utilize intricate inter-space and intra-space alignment loss to train their well-designed projector. Their tasks aims to align two expert spaces with one and only one shared modality, and the intra-space alignment loss is used to better transfer the robust connections learned from the shared modality to non-shared modalities.

In contrast, our objective is to enhance a pre-trained unified space by integrating expert spaces. Given unified space typically covers most modality inputs, and the modalities of expert spaces are the subset of unified space. The space alignment learned from the multiple shared modalities is much stronger than that learned from only one shared modality. Therefore, there is no motivation for using intra-space alignment loss here, and previous complex learning pipeline may introduce a negative impact on generalization.

As a result, we propose a more plain space alignment pipeline, which experimentally shows better performance. One projector $\psi_i$ consists of simple multi-layer perceptrons (MLP). For the learning objective, we only compute the InfoNCE loss, denoted as $\operatorname{info}(\cdot,\cdot)$, between features of different spaces. The training loss for displacement bond in Figure~\ref{fig:model} can be expressed as:
\begin{equation}
\label{eq:dr_loss}
    \begin{aligned}
        L & = \operatorname{info}(\tilde{\mathbf{T}}^{vt},\psi_i^{u}(\tilde{\mathbf{T}}^u)) + \operatorname{info}(\tilde{\mathbf{T}}^{vt},\psi_i^{u}(\tilde{\mathbf{V}}^u)) \\ &  + \operatorname{info}
        (\tilde{\mathbf{T}}^{vt},\psi_i^{u}(\tilde{\mathbf{A}}^u)) + \operatorname{info}(\tilde{\mathbf{V}}^{vt},\psi_i^{u}(\tilde{\mathbf{T}}^u)) \\ & + \operatorname{info}(\tilde{\mathbf{V}}^{vt},\psi_i^{u}(\tilde{\mathbf{V}}^u)) + \operatorname{info}(\tilde{\mathbf{V}}^{vt},\psi_i^{u}(\tilde{\mathbf{A}}^u))
    \end{aligned}
\end{equation}
and the loss for the combination bond in Figure~\ref{fig:model} is:
\begin{equation}
\label{eq:cr_loss}
    \begin{aligned}
        L & = \operatorname{info}(\psi_i^{at}(\tilde{\mathbf{T}}^{at}),\tilde{\mathbf{T}}^u) + \operatorname{info}(\psi_i^{at}(\tilde{\mathbf{T}}^{at}),\tilde{\mathbf{V}}^u) \\ &  + \operatorname{info}
        (\psi_i^{at}(\tilde{\mathbf{T}}^{at}),\tilde{\mathbf{A}}^u) + \operatorname{info}(\psi_i^{at}(\tilde{\mathbf{V}}^{at}),\tilde{\mathbf{T}}^u) \\ & + \operatorname{info}(\psi_i^{at}(\tilde{\mathbf{V}}^{at}),\tilde{\mathbf{V}}^u) + \operatorname{info}(\psi_i^{at}(\tilde{\mathbf{V}}^{at}),\tilde{\mathbf{A}}^u)
    \end{aligned}
\end{equation}

\paragraph{Mixture-of-Projectors Strategy}
Inspired by the ensemble learning and mixture-of-expert methods, we propose the mixture-of-projectors strategy, which learns multiple projectors with different training data and ensembles them to achieve more robust alignment and more discriminative representations. Specifically, we first sample $t$ subsets from the whole dataset $D$, denoted as $\{ {D}_1, {D}_2, \ldots, {D}_t \}$. Then we train projector $\psi_i$ on ${D}_i$ respectively, and finally get a group of projectors $\Psi = \{ \psi_1, \psi_2, \ldots, \psi_t \}$. The output of $\Psi$ is the mean pool of all $t$ projectors.

\subsubsection{Inference}
In the product space, one modality may have multiple representations from different sources. As illustrated in parts 1.3 and 2.3 in Figure~\ref{fig:model}, we simply weighted average the representations of the same modality but from different sources.


\subsection{Complex Sequential \& Parallel Bonds}
\label{sec:complex}
Based on these two basic bonds, we can easily construct various complex bonds, but which way is more effective for integrating multiple spaces still needs to be explored.




Typical unified space learning method aligns encoders of other modalities to pre-trained image-text space via massive paired data. Therefore, image-text representation is the foundation of unified spaces and directly determines its potential. Considering the properties of basic bonds and the importance of image-text space, we propose sequential \& parallel bonds, which consist of two stages:

1) Sequential Displacement. Given the pivotal role of image-text representation and the value of image-text knowledge (requiring training encoders of billion-level parameters on billion-level data), we integrate advanced image-text space via displacement bond and tuning on data of other modalities to repair the missing knowledge of unified space. 

2) Parallel Combination. After obtaining stronger image-text representations, we integrate expert spaces of other modalities in parallel via combination bonds. Since these expert spaces are independently connected to the same frozen unified space, we can further enhance the unified space and perform flexible customized inference.


Take the integration of advanced image-text space and $n$ audio-text spaces as an example. Based on the displacement product, $\hat{\mathcal{A}}^u(\mathcal{\hat{V}}^u_{1\!-\!\lambda_v}\mathcal{V}^{vt}_{\lambda_v})(\hat{\mathcal{T}}^u_{1\!-\!\lambda_t}\mathcal{T}^{vt}_{\lambda_t})$, the combination bond of the $i$-th audio-text space can be formulated as:
\begin{equation}
\begin{aligned}
    \label{complex reaction}
     \hat{\mathcal{A}}^u(\mathcal{\hat{V}}^u_{1\!-\!\lambda_v}&\mathcal{V}^{vt}_{\lambda_v})(\hat{\mathcal{T}}^u_{1\!-\!\lambda_t}\mathcal{T}^{vt}_{\lambda_t}) + \operatorname{c}(\mathcal{A}^{at_i}\mathcal{T}^{at_i}){\to}\\
    & (\hat{\mathcal{A}}_{1\!-\!\sigma\!_a}^u\hat{\mathcal{A}}^{at_i}_{\sigma\!_a})(\mathcal{\hat{V}}_{1\!-\!\lambda_v}^u\mathcal{V}^{vt}_{\lambda_v})[(\hat{\mathcal{T}}^u_{1\!-\!\lambda_t}\mathcal{T}^{vt}_{\lambda_t})_{1\!-\!\sigma\!_{t}}\hat{\mathcal{T}}^{at_i}_{\sigma\!_{t}}]\\
\end{aligned}
\end{equation}

Since $n$ audio-text spaces are aligned to the same unified space, $\hat{\mathcal{A}}^{at}_i$ and $\hat{\mathcal{T}}^{at}_i$ can be flexibly combined during inference to obtain customized representations. 
The space combined all the $n$ audio-text space can be formulated as:  $(\hat{\mathcal{A}}_{1\!-\!\sigma\!_a}^u\frac{1}{n}\!\sum\limits_{i=1}^n\!\hat{\mathcal{A}}^{at_i}_{\sigma\!_a})(\mathcal{\hat{V}}_{1\!-\!\lambda_v}^u\mathcal{V}^{vt}_{\lambda_v})[(\hat{\mathcal{T}}^u_{1\!-\!\lambda_t}\mathcal{T}^{vt}_{\lambda_t})_{1\!-\!\sigma\!_{t}}\frac{1}{n}\!\sum\limits_{i=1}^n\!\hat{\mathcal{T}}^{at_i}_{\sigma\!_{t}}]$, and its combining process can be expressed as:
\begin{equation}
\label{eq:complex_inference}
    \begin{aligned} 
    (\hat{\mathcal{A}}_{1\!-\!\sigma\!_a}^u\frac{1}{n}\!\sum\limits_{i=1}^n\!\hat{\mathcal{A}}^{at_i}_{\sigma\!_a})\! &= ({1\!-\!\sigma\!_a})\hat{\mathcal{A}}^u \!+\! \frac{\sigma\!_a}{n}\sum_{i=1}^n\hat{\mathcal{A}}^{at_i}; \\
    [(\hat{\mathcal{T}}^u_{1\!-\!\lambda_t}\mathcal{T}^{vt}_{\lambda_t})_{1\!-\!\sigma\!_{t}}\frac{1}{n}\!\sum\limits_{i=1}^n\!\hat{\mathcal{T}}^{at_i}_{\sigma\!_{t}}]\! &=\! (1\!-\!\sigma\!_{t}) (\hat{\mathcal{T}}^u_{1\!-\!\lambda_t}\mathcal{T}^{vt}_{\lambda_t}) \!+\! \frac{\sigma\!_t}{n}\sum_{i=1}^n\hat{\mathcal{T}}^{at_i}
    \end{aligned}
\end{equation}

\subsection{Coarse-to-Fine Customized Inference}
\label{sec:customized}
In addition to the computationally efficient training process, the product of FreeBind can customize its inference to various applications. To fully realize its potential, we propose a coarse-to-fine customized inference strategy:


1) Coarse-grained Combined Modules Selection. Combination bonds align multiple expert spaces into a unified space. Therefore, during inference, we can flexibly select any aligned expert spaces to obtain gains of specific aspects.

2) Fine-grained Combining Factors Adjustment. In addition to selecting different modules, we can also customize the enhanced unified space in a fine-grained manner by changing the combination weights of different expert spaces. 

Using the inference process in Equation~\ref{eq:complex_inference} as an example, we can freely select any combination of the $n$ aligned audio-text spaces to construct unified spaces tailored to specific aspects. Additionally, a small ($\sigma\!_{a}, \sigma\!_{t}$) implies partial absorption of audio-text knowledge, and moderate knowledge fusion can enhance both audio-text and audio-image performance while maintain advanced image-text ability. Conversely, a larger value for ($\sigma\!_{a}, \sigma\!_{t}$) leads to superior audio-text performance at the expense of other alignments. Notably, the impact of combination factors on performance is regular and robust. As depicted in Figure~\ref{fig:analysis}, most settings either yield a versatile space that surpasses the original unified space or generate an expertise space that exceeds the source expert spaces in their fields.



\section{Experiment and Discussions}
\subsection{Implementation Details}
\paragraph{Data and Pre-trained Models} For both bonds, we employ 2.3M unpaired texts, 1.3M images, and 1.8M audios, following~\cite{wang2023connecting}. We optionally use the audio-image pairs in AudioSet~\cite{gemmeke2017audio} (the audio pre-training dataset of ImageBind) to fine-tune the audio encoder. We enhance the unified audio-image-text space of ImageBind by integrating one image-text expert space, InternVL-C~\cite{chen2023internvl} and two audio-text expert spaces, two versions of CLAPs~\cite{wu2023large}.
\paragraph{Training and Inference} For both kinds of basic bond, the temperature of softmax in data collection is 1/100, and the temperature of InfoNCE loss is 1/50. We leverage the all possible combination of the elements $D_T$, $D_V$ and $D_A$ as the sampled subsets in mixture-of-projector (i.e., $D_T$, $D_V$, $D_A$, $D_{TV}$, $D_{T\!A}$, $D_{V\!A}$, $D_{TV\!A}$), resulting in 7 projectors of each $\Psi$. 
All our experiments are conducted on a single 4090 GPU. We use Adam~\cite{kingma2014adam} optimizer with a learning rate of 1e-3 and batch size of 4096 for both bond. The displacement bond is trained for 5 epochs, while the combination bond is trained for 20 epochs.
\paragraph{Evaluation Protocols} We comprehensively evaluate FreeBind on nine datasets over five zero-shot downstream tasks. The evaluation tasks, datasets, metrics, and the number of test samples are summarized in Table~\ref{tab:dataset}.

\begin{table}[]
\centering
\vspace{-0.6\baselineskip}
\caption{Summary of downstream tasks and datasets.}
\setlength\tabcolsep{3pt}
\label{tab:dataset}
\renewcommand{\arraystretch}{1}
\resizebox{0.48\textwidth}{!}{
\begin{tabular}{c|ccc}
\toprule
\textbf{Task} & \textbf{Dataset} & \textbf{Metric} & \textbf{\#Samples} \\ \midrule
\multirow{2}{*}{\begin{tabular}[c]{@{}c@{}}Audio-Text\\ Retrieval\end{tabular}} & AudioCaps~\cite{kim2019audiocaps} & Recall & 964 \\
 & Clotho~\cite{drossos2020clotho} & Recall & 1045 \\ \midrule
\multirow{2}{*}{\begin{tabular}[c]{@{}c@{}}Audio-Image\\ Retrieval\end{tabular}} & VGG-SS~\cite{chen2021localizing} & Recall & 5158 \\
 & FlickrNet~\cite{senocak2018learning} & Recall & 5000 \\ \midrule
\multirow{2}{*}{\begin{tabular}[c]{@{}c@{}}Image-Text\\ Retrieval\end{tabular}} & COCO~\cite{lin2014microsoft} & Recall & 5000 \\
 & Flickr30k~\cite{young2014image} & Recall & 1000 \\ \midrule
\multirow{2}{*}{\begin{tabular}[c]{@{}c@{}}Audio\\ Classification\end{tabular}} & ESC-50~\cite{piczak2015esc} & Acc & 400 \\
 & AudioSet~\cite{gemmeke2017audio} & mAP & 19048 \\ \midrule
\multirow{2}{*}{\begin{tabular}[c]{@{}c@{}}Image\\ Classification\end{tabular}} & \multirow{2}{*}{ImageNet~\cite{deng2009imagenet}} & \multirow{2}{*}{Acc} & \multirow{2}{*}{50000} \\
 &  &  &  \\ \bottomrule
\end{tabular}}
\end{table}

\begin{table}[t]
\vspace{-1\baselineskip}
\caption{Notations of different bond processes and their corresponding resulting space.}
\centering
\setlength\tabcolsep{5pt}
\label{tab:notation}
\renewcommand{\arraystretch}{1}
\resizebox{0.48\textwidth}{!}{
\begin{tabular}{lc}
\toprule
\textbf{Bonds Combination} & \textbf{Product} \\ \midrule
ImageBind +$\operatorname{c}$(CLAP$_{m}$)+$\operatorname{c}$(CLAP$_{g}$) & ImageBind++ \\
ImageBind+$\operatorname{d}$(InternVL) & InternVL$_{I\!B}$ \\
InternVL$_{I\!B}$+$\operatorname{c}$(CLAP$_{m}$) + $\operatorname{c}$(CLAP$_{g}$) & InternVL$_{I\!B}$++ \\
InternVL$^\dagger_{I\!B}$+$\operatorname{c}$(CLAP$_{m}$)+$\operatorname{c}$(CLAP$_{g}$) & InternVL$^\dagger_{I\!B}$++ \\ \bottomrule
\end{tabular}}
\vspace{-1\baselineskip}
\end{table}

\begin{table}[t]
\vspace{-1\baselineskip}
\caption{Results of zero-shot classification.}
\centering
\setlength\tabcolsep{3pt}
\label{tab:cls}
\renewcommand{\arraystretch}{1}
\resizebox{0.48\textwidth}{!}{
\begin{tabular}{c|l|cc|cc}
\toprule
& \multirow{2}{*}{Models} & \multicolumn{2}{c|}{Audio} & \multicolumn{1}{c}{Image} \\
&  & ESC-50 & AudioSet & ImageNet \\ \midrule
\multirow{3}{*}{\begin{tabular}[c]{@{}c@{}}Pre-trained \\ Expert Space\end{tabular}} & InternVL & -&	-&	81.70   \\
& CLAP$_{g}$ & 90.95 &	23.36 &	-  \\
& CLAP$_{m}$ &92.60 &	23.08 &	-  \\ \midrule
\multirow{5}{*}{\begin{tabular}[c]{@{}c@{}}Pre-trained \\ Unified Space\end{tabular}} & WAV2CLIP & 17.40 &	0.71 &	60.58   \\
& AudioCLIP &11.45 &	5.65 &	24.14   \\
& C-MCR &  66.05 &	11.15 &	23.16   \\
& Ex-MCR & 68.35 &	6.67 &	60.58   \\
& ImageBind & 67.25 &	13.96 &	76.31   \\ \midrule
\multirow{8}{*}{\begin{tabular}[c]{@{}c@{}}Enhanced \\ Unified Space\end{tabular}} &  InternVL$_{I\!B}$ & 66.05 &	11.82 &	\underline{81.54}   \\
&  InternVL$^{\dagger}_{I\!B}$ & 66.75 &	11.82 &	\textbf{81.54}   \\ \cmidrule{2-5}

& ImageBind++(\textit{AT E.})& 92.80 &	25.35 &	74.83   \\
& InternVL$_{I\!B}$++(\textit{AT E.})& \textbf{93.60} 	&\underline{25.35} &	73.33    \\
& InternVL$^\dagger_{I\!B}$++(\textit{AT E.}) & \underline{93.00} &	\textbf{26.45} &	72.07    \\ \cmidrule{2-5}

&  ImageBind++(\textit{Ver.})& 88.55 	&19.69 &	76.32   \\
&  InternVL$_{I\!B}$++(\textit{Ver.}) & 88.30 &	18.93 &	81.49    \\
&  InternVL$^{\dagger}_{I\!B}$++(\textit{Ver.}) & 87.70 &	19.23 &	81.42    \\
 \bottomrule
\end{tabular}}
\vspace{-1\baselineskip}
\end{table}

\begin{table*}[]
\vspace{-1\baselineskip}
\caption{Results of zero-shot cross-modal retrievals. The best result is \textbf{bolded}, and the second best result is \underline{underlined}.}
\centering
\setlength\tabcolsep{5pt}
\label{tab:retrieval}
\renewcommand{\arraystretch}{1}
\resizebox{1\textwidth}{!}{
\begin{tabular}{c|l|cccc|cccc|cccc}
\toprule
\multirow{3}{*}{} & \multirow{3}{*}{Models} & \multicolumn{4}{c|}{Audio-Text} & \multicolumn{4}{c|}{Audio-Image} & \multicolumn{4}{c}{Image-Text} \\
 &  & \multicolumn{2}{c}{AudioCaps} & \multicolumn{2}{c|}{Clotho} & \multicolumn{2}{c}{VGG-SS} & \multicolumn{2}{c|}{FlickrNet} & \multicolumn{2}{c}{COCO} & \multicolumn{2}{c}{Flickr30K} \\
 &  & R@1 & R@5 & R@1 & R@5 & R@1 & R@5 & R@1 & R@5 & R@1 & R@5 & R@1 & R@5 \\ \midrule
\multirow{3}{*}{\begin{tabular}[c]{@{}c@{}}Pre-trained \\ Expert Space\end{tabular}} & InternVL &    -& 	-& 	-& 	-& 	-& 	-& 	-& 	-& 	61.05& 	82.18&  	89.29&  	98.19  \\
& CLAP$_{g}$ &  40.25 &	 76.21 &	 18.46 &	 41.70 &	 -	& -&	 -&	 -	& -	& -	& -	& - \\
 & CLAP$_{m}$ &  39.00 &	 74.71 &	 17.90 &	 40.57 &	 -	& -	& -	& -&	 -	& -	& -	& - \\ \midrule
\multirow{5}{*}{\begin{tabular}[c]{@{}c@{}}Pre-trained \\ Unified Space\end{tabular}} & WAV2CLIP & 0.88 &	4.22 &	0.78 &	2.60 &	2.51 &	9.28 &	0.82 &	3.41 &	40.24 &	64.78 &	71.89 &	90.55   \\
 & AudioCLIP &3.53 &	11.30 &	3.20 &	9.97 &	1.25 &	3.91 &	1.37 &	4.91 &	17.51 &37.50 &	38.89 &	64.92  \\
 & C-MCR & 15.76 &	41.37 &	8.37 &	24.86 &	1.94 &	7.69 &	1.39 &	5.97 &	16.67 &	37.04 &	34.16 &	63.64  \\
 & Ex-MCR & 19.07 &	47.05 &	7.01 &	22.04 &	2.13 &	8.13 &	1.57 &	5.94& 	40.24 &	64.78 &	71.89 &	90.55  \\
 & ImageBind & 9.24 &	27.47 &	6.64 &	17.28 &	14.82 &	35.67 &	7.68 &	20.79 &	57.28 &	79.54 &	86.04 &	96.97    \\ \midrule
\multirow{8}{*}{\begin{tabular}[c]{@{}c@{}}Enhanced \\ Unified Space\end{tabular}}
 & InternVL$_{I\!B}$ & 11.17 &	32.03 &	6.60 &	18.27 &	13.77 	&34.56 &	7.37 &	20.80 &	\underline{61.49} &	\underline{82.17} 	&\underline{89.61} &	\underline{98.14}   \\
 & InternVL$_{I\!B}^{\dagger}$ & 11.20 &	31.84 &	6.42 &	17.45 	&14.47 &	36.44 &	\underline{8.02}& 	20.90 	&\textbf{61.49} &	\textbf{82.17} &	\textbf{89.61} &	\textbf{98.14}   \\ \cmidrule{2-14} 

& ImageBind++(\textit{AT E.})& \underline{42.50} &	77.31 &	\underline{19.49} &	43.57 	&12.38 &	33.47& 	5.63 &	17.35 &	47.61 &	71.82 &	78.52 &	94.37   \\
& InternVL$_{I\!B}$++(\textit{AT E.})& 41.41 &	\underline{77.68} &	\textbf{20.19} 	&\textbf{46.03} &	10.99 &	30.62 &	5.24 &	17.65 &	54.87 &	78.00 &	85.45 	&96.81   \\
& InternVL$^\dagger_{I\!B}$++(\textit{AT E.}) & \textbf{43.10} &	\textbf{78.31} 	&19.38 	&\underline{45.17} &	12.23 &	32.68 &	5.71 &	18.23& 	54.75 &	77.87 &	85.08 &	96.99   \\ \cmidrule{2-14} 

& ImageBind++(\textit{Ver.})& 29.16 &	62.98 &	13.67 &	33.19 	&\underline{15.48} &	\textbf{39.26} &	8.01 &	\underline{21.87} &	57.01 &	79.23  &	85.91  &	97.03   \\
& InternVL$_{I\!B}$++(\textit{Ver.}) & 29.11  &	62.30 &	12.66 &	32.75 &	14.40 &	36.78 &	7.74 &	21.85 &	61.07& 	82.00 &	89.30 &	98.09  \\
& InternVL$^\dagger_{I\!B}$++(\textit{Ver.}) &  29.42 &	62.10 &	13.48 &	33.69 &	\textbf{15.48} &	39.14 &	\textbf{8.32} &	\textbf{22.40} &	61.13 &	82.05 &	89.33 	&98.11    \\
 \bottomrule
 \end{tabular}}
 \vspace{-1.5\baselineskip}
\end{table*}

\subsection{Augmenting ImageBind}
To show the effectiveness of the proposed methods, we augment the audio-image-text space of ImageBind with InternVL (image-text space), CLAP$_{g}$ (audio-text space for general purpose), and CLAP$_{m}$ (audio-text space for music purpose). For simplicity of expression, we summarize the notations of output space for different bonds combination in Table~\ref{tab:notation}. The InternVL$_{I\!B}$ tuned on the audio-image dataset is denoted as InternVL$^\dagger_{I\!B}$. There are two standard settings of combining factors: Versatile (\textit{Ver.}) and Audio-Text Expertise (\textit{AT E.})\footnote[1]{The CLAPs combining factors ($\sigma\!_a$, $\sigma\!_t$) are (0.5, 0.1) for Versatile (\textit{Ver.}) and (0.8, 0.5) for Audio-Text Expertise (\textit{AT E.}). The InternVL combining factors ($\lambda_v$, $\lambda_t$) in InternVL$_{I\!B}$ are (0.9, 0.9).}. The zero-shot classification results are presented in Table~\ref{tab:cls}, and the multimodal retrieval results can be found in Table~\ref{tab:retrieval}.
\paragraph{Displacement Bond}
By integrating InternVL with ImageBind via displacement bond, the resulting unified space InternVL$_{I\!B}$ shows significantly better image-text performance than ImageBind. Additionally, its image-text retrieval accuracy even surpasses the source image-text space, InternVL. More importantly, despite the audio representation in InternVL$_{I\!B}$ is a remapped and degraded version of ImageBind's audio representations, InternVL$_{I\!B}$ achieves comparable audio-text and audio-image performance.


\paragraph{Combination Bond} 
We try to integrate CLAPs for three kinds of unified space: ImageBind, InternVL$_{I\!B}$ and InternVL$_{I\!B}^\dagger$. The products are denoted as ImageBind++, InternVL$_{I\!B}$++ and InternVL$_{I\!B}^\dagger$++, respectively. 

The (\textit{Ver.}) variants achieve much better audio classification and audio-text retrieval performance than their corresponding source unified space while maintaining the image-text capabilities of source space. More importantly, although the audio representation in CLAP is learned by aligning with text, absorbing it can even improve audio-image alignment in the unified space. This discovery highlights cross-modal knowledge transfer capabilities, further broadening the potential of knowledge fusion in multimodal representations. In summary, combining expert spaces with appropriate factor can significantly enhance corresponding aspects without incurring extra costs, akin to a free lunch.



Additionally, increasing CLAP's combining weights yields an audio-text expertise unified space, denoted as (\textit{AT E.}). This variant achieves even better audio-text retrieval and audio classification accuracy than CLAPs, while maintaining competitive performance for other multimodal alignments.

\paragraph{Complex Sequential \& Parallel Bonds} 
As a result of our complex sequential \& parallel bonds, InternVL$^\dagger_{I\!B}$++ exhibits a significant advantage in image-text fields compared to ImageBind++, while achieving similar state-of-the-art performance in audio-related tasks. Besides, the overall advantage of InternVL$^\dagger_{I\!B}$++ over InternVL$_{I\!B}$++ demonstrates that simply tuning the small audio encoder with limited resources can effectively repair the lost knowledge.

Notably, considering the pivotal role of image-text representation in unified space, tuning the image or text encoder not only demands massive computing resources but also potentially compromises the foundation of the unified space. Therefore, repairing lost knowledge through fine-tuning is only suitable for modalities other than image or text, which further emphasizes the essential of preserving the advanced image-text expert knowledge.

\begin{table}[]
\caption{\textbf{Study of Basic Displacement and Combination Bond.} The reported retrieval metric is R@1. ``ACaps" stands for AudioCaps. The combining factors are set to (0.9, 0.9).}
\centering
\setlength\tabcolsep{2pt}
\label{tab:reaction}
\renewcommand{\arraystretch}{1}
\resizebox{0.48\textwidth}{!}{
\begin{tabular}{l|cc|cc|cc}
\toprule
\multirow{2}{*}{Setting} & \multicolumn{2}{c|}{Audio-Text} & \multicolumn{2}{c|}{Audio-Image} &\multicolumn{2}{c}{Image-Text} \\
 & ACaps & Clotho & VGGSS & FlickrNet & COCO & Flickr30K \\ \midrule
ImageBind & 9.24 &	6.64 &	14.82 &	7.68 &	57.28 &	86.04   \\
+$\operatorname{Cr}$(InternVL) &9.25 &	\textbf{6.64} &	\textbf{16.14} &	\textbf{8.08} &	58.90 &	87.93 \\
+$\operatorname{Dr}$(InternVL) & \textbf{11.17} &	6.60 &	13.77 &	7.37 &	\textbf{61.49} &	\textbf{89.61} \\ \bottomrule
\end{tabular}}
\vspace{-1\baselineskip}
\end{table}

\begin{table}[t]
\caption{\textbf{Study of Different Complex Bonds.} ``Only seq" means sequentially integrating InternVL, CLAP$_{m}$, and CLAP$_{g}$ via displacement. ``Only Para." means aligning in parallel with combination. ``Seq.\&Para." refers to our complex sequential \& parallel bonds. The combining factors follow (\textit{Ver.}) variant.}
\centering
\setlength\tabcolsep{2pt}
\label{tab:route}
\renewcommand{\arraystretch}{1}
\resizebox{0.48\textwidth}{!}{
\begin{tabular}{c|cc|cc|cc}
\toprule
\multirow{2}{*}{Route} & \multicolumn{2}{c|}{Audio-Text} & \multicolumn{2}{c|}{Audio-Image} &\multicolumn{2}{c}{Image-Text} \\
 & ACaps & Clotho & VGGSS & FlickrNet & COCO & Flickr30K \\ \midrule
Only Seq. & \textbf{36.03} &	\textbf{16.62} &	9.24 &	4.89 &	33.77 	& 66.55  \\
Only Para.\tablefootnote[2]{Given InternVL are remapped, its combining factor is set to (0.1, 0.1), which corresponds to (0.9, 0.9) in InternVL$_{I\!B}$.} & 29.14 &	13.48 &	\textbf{16.06} &	8.20 &	58.83 &	87.35   \\
Seq.\&Para. & 29.42 &	13.48 &	15.48 &	\textbf{8.32} &	\textbf{61.13} &	\textbf{89.33}  \\ \bottomrule
\end{tabular}}
\vspace{-1.2\baselineskip}
\end{table}

\subsection{Discussion}
\label{sec:disc}
\paragraph{Displacement or Combination} We integrate InternVL and ImageBind through two basic bonds to further reveal their properties. As shown in Table~\ref{tab:reaction}, the displacement bond inherits InternVL's advanced image-text capabilities. Despite audio embeddings are re-projected, the resulting space still achieves comparable performance in audio-text and audio-image retrieval. Meanwhile, the combination bond yield slight but consistent improvements over ImageBind. These observations reinforce our analysis of the basic bonds: the displacement bond is a radical knowledge fusion scheme, whereas the combination bond is more moderate.

\begin{table}[t]
\caption{\textbf{Study of Corase-grain Module Selection.} The combining factors follow the (\textit{Ver.}) variant.}
\centering
\setlength\tabcolsep{3pt}
\label{tab:corase}
\renewcommand{\arraystretch}{1}
\resizebox{0.48\textwidth}{!}{
\begin{tabular}{l|cc|ccc}
\toprule
\multirow{2}{*}{Setting} & \multicolumn{2}{c|}{Classification} &  \multicolumn{2}{c}{Retrieval} \\
 & ESC & AudioSet & ACaps & Clotho \\ \midrule
ImageBind & 67.25 &	13.96 & 9.24 & 	6.64  \\
+$\operatorname{Cr}$(CLAP$_g$) &86.65 &	19.09& 28.63 &	13.24   \\
+$\operatorname{Cr}$(CLAP$_m$) &88.40 &	18.89 & 27.33 &	12.70 \\
+$\operatorname{Cr}$(CLAP$_g$)+$\operatorname{Cr}$(CLAP$_m$) & \textbf{88.55} &	\textbf{19.69} & \textbf{29.16} & \textbf{13.67}   \\ \bottomrule
\end{tabular}}
\vspace{-0.8\baselineskip}
\end{table}

\paragraph{Different Complex Bonds} We compare different complex bonds in Table~\ref{tab:route}. Our complex sequential \& parallel bonds achieve more balanced and stable improvements than pure sequential or parallel routes that rely on only one bond. These results confirm our analysis in Section~\ref{sec:complex} and emphasize the importance of combining the complementary basic bonds when designing complex bonds.

\paragraph{Corase-grain Combined Module Selection} Table~\ref{tab:corase} report the results of employing different aligned audio-text experts to enhance ImageBind. The results reveal that combining different modules exhibits varying abilities. Integrating CLAP$_{m}$ yields more gains on ESC, while CLAP$_{g}$ improves performance more on other general audio datasets. Employing both together brings better overall performance.

\begin{figure}[t]
	\centering
	\includegraphics[width=\linewidth]{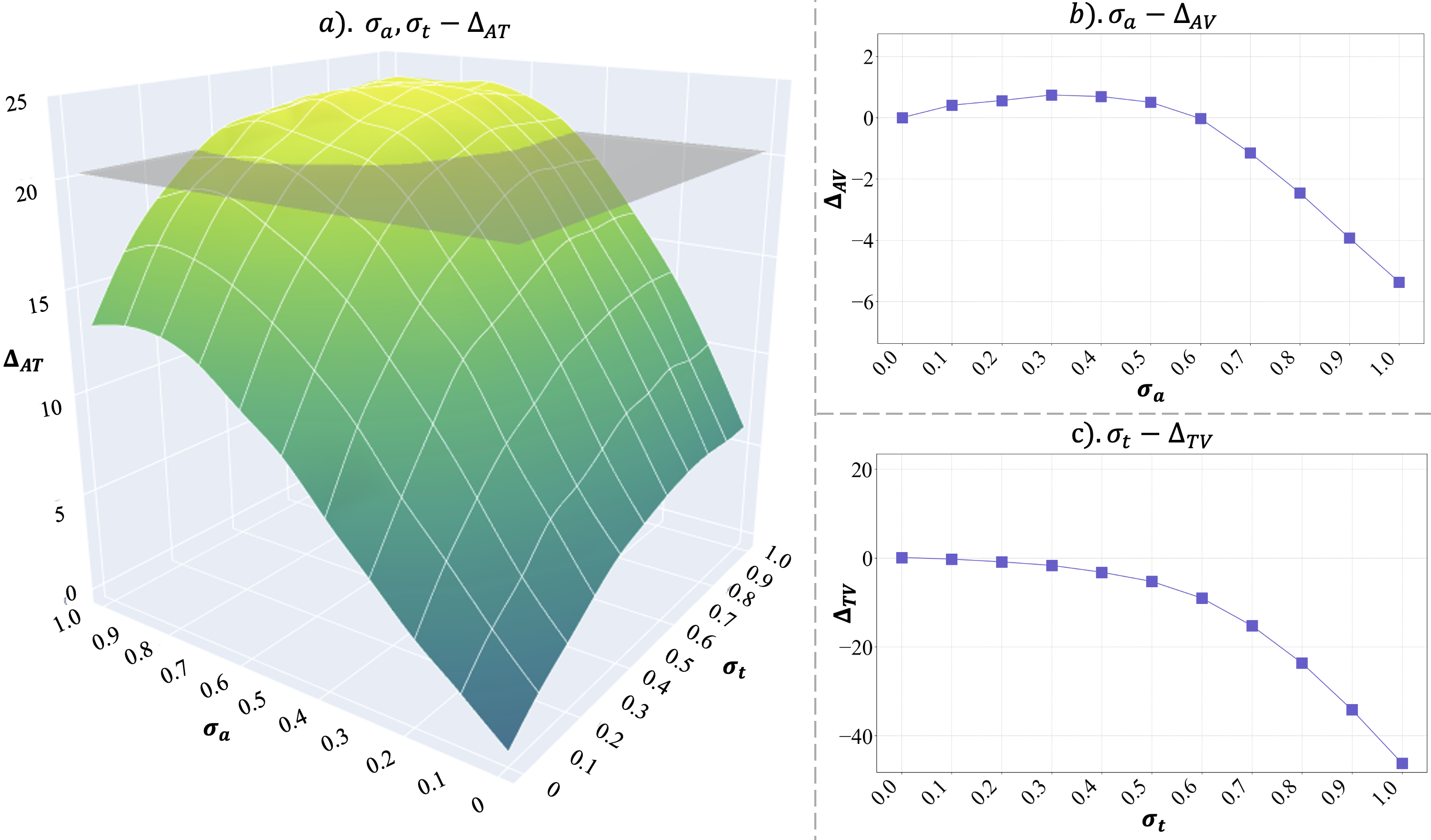}
        \vspace{-1\baselineskip}
	\caption{\textbf{Analysis of CLAPs' combining factors ($\sigma_a, \sigma_t$) on InternVL$_{I\!B}^\dagger$++.} $\Delta_{AT}, \Delta_{A\!V}, \Delta_{TV}$ represents the average R@1 variance between InternVL$_{I\!B}^\dagger$++ and InternVL$_{I\!B}^\dagger$ on audio-text, audio-image and image-text retrieval tasks, respectively. Positive $\Delta_{*}$ signifies improvements in the corresponding task, while negative values indicate reductions. The gray plane in the 3D figure $a)$ denotes the audio-text performance of CLAP$_{g}$. }
        \vspace{-\baselineskip}
        \label{fig:analysis}
\end{figure}
\paragraph{Fine-grained Combining Factors Adjustment} To explore and provide insights about combining factors adjustment, we comprehensively display the effect of ($\sigma\!_a$, $\sigma\!_t$) on InternVL$_{I\!B}^\dagger$ in Figure~\ref{fig:analysis}. There are three main observations: 1) All ($\sigma\!_a$, $\sigma\!_t$) can significantly enhance audio-text alignment, and when they are set larger than 0.5, the enhanced unified space even outperforms CLAP$_g$ in audio-text field. 2) When $\sigma\!_a$ takes a moderate value (around 0.5), the audio-image performance can be improved. 3) Since CLAP's text representation is aligned to audio, large $\sigma\!_t$ may hurt the image-text alignment, but when it is set to small value (around 0.2), the negative effect is negligible. 

Generally speaking, the combining factors adjustment are logical and insensitive. Most settings either bring an overall stronger unified space or provide superior expertise in a certain aspect. We conduct more analyses and visualizations on all the resulting space in the Appendix~\ref{sec:further}, which further prove the regularity and insensitivity of combining factors.

\paragraph{Mixture-of-Projectors} Results in Table~\ref{tab:mop} illustrate that combining all projectors yields substantial performance benefits, which prove that our mixture-of-projectors strategy enhances alignment and fosters more discriminative representations. Noteworthy, each projector typically consists of about 2M parameters, therefore multiple projectors will only incur minimal extra inference costs.

\paragraph{Projector Design} We investigate various projector structures and learning objective designs, and the results are reported in Table~\ref{tab:proj}. Compared with the projector learning pipeline proposed in the previous advanced space integration methods C-MCR and Ex-MCR, our simpler pipeline achieves better overall results in both basic bonds. The multiple shared modalities between unified space and expert spaces can sufficiently align the spaces. In this scenario, complex learning pipelines and intra-space loss may hinder alignment generalization. Our straightforward design is better suited for unified space scenarios.



\begin{table}[t]
\caption{\textbf{Study of Mixture-of-Projector.} The results are obtained on bond: ImageBind+$\operatorname{Dr}$(InternVL). $\psi_{*}$ represents the projector trained on subset $D_{*}$. $\Psi$ indicates the whole projectors group. } 
\centering
\setlength\tabcolsep{3pt}
\label{tab:mop}
\renewcommand{\arraystretch}{1}
\resizebox{0.48\textwidth}{!}{
\begin{tabular}{c|cc|cc|cc}
\toprule
\multirow{2}{*}{\begin{tabular}[c]{@{}c@{}}Used \\ Projector\end{tabular}} & \multicolumn{2}{c|}{Audio-Text} & \multicolumn{2}{c|}{Audio-Image} &\multicolumn{2}{c}{Image-Text} \\
 & ACaps & Clotho & VGGSS & FlickrNet & COCO & Flickr30K \\ \midrule
 $\psi_A$ &9.59 &	5.79 &	12.61 &	6.39 &	61.44 & 89.10 \\
 $\psi_V$ & 9.79 &	5.53 &	10.83 &	5.97 &	61.30  & 89.32 \\
 $\psi_T$ & 10.07 &5.57 &	10.08 &	6.17 &	61.47  & 89.24 \\
 $\psi_{AT}$ & 9.96 &	5.86 &	12.60 &	6.68 &	61.45 & 89.26 \\
 $\psi_{V\!A}$ &9.77 &	5.91 &	12.93 &	6.74 &	61.44 & 89.33 \\
 $\psi_{VT}$ & 9.91 &	5.46 &	11.12 &	6.43 &	61.49 & 89.24 \\
 $\psi_{V\!AT}$ & 10.23 &	5.98 &	12.70 &	6.78 &	61.42  & 89.30 \\ \midrule
 $\Psi$ & \textbf{11.17} &	\textbf{6.60} &	\textbf{13.77} &	\textbf{7.37} &	\textbf{61.49} & \textbf{89.61} \\ \bottomrule
\end{tabular}}
\end{table}

\begin{table}[]
\vspace{-1\baselineskip}
\caption{\textbf{Study of Projector Design.} The projector structure and learning objective of C-MCR, Ex-MCR and Ours are used for two basic bonds, respectively.}
\centering
\setlength\tabcolsep{2pt}
\label{tab:proj}
\renewcommand{\arraystretch}{1}
\resizebox{0.48\textwidth}{!}{
\begin{tabular}{c|cc|cc|cc}
\toprule
\multirow{2}{*}{\begin{tabular}[c]{@{}c@{}}Training \\ Pipeline\end{tabular}} & \multicolumn{2}{c|}{Audio-Text} & \multicolumn{2}{c|}{Audio-Image} &\multicolumn{2}{c}{Image-Text} \\
 & ACaps & Clotho & VGGSS & FlickrNet & COCO & Flickr30K \\ \midrule
 \multicolumn{7}{c}{\footnotesize Combination Bond --- IB + $\operatorname{Cr}$(InternVL)} \\ \midrule
 C-MCR &\textbf{9.26} &	6.64 &	15.99 &	\textbf{8.12} &	58.62 &	87.77  \\
 Ex-MCR & 9.25 &	6.64 &	15.60 &	7.65 &	58.47 &	87.11  \\
 Ours & 9.25 &	\textbf{6.64} &	\textbf{16.14} &	8.08 &	\textbf{58.90} &	\textbf{87.93}  \\ \midrule
 \multicolumn{7}{c}{\footnotesize Displacement Bond --- IB + $\operatorname{Dr}$(InternVL)} \\ \midrule
 C-MCR & 10.43 &	5.55 &	11.60 &	6.32 &	61.28 &	89.13  \\
 Ex-MCR & 8.97 &	5.27 &	8.78 &	5.09 &	61.26 &	89.48 \\
 Ours &  \textbf{11.17} &	\textbf{6.60} &	\textbf{13.77} &	\textbf{7.37} &	\textbf{61.49} &	\textbf{89.61} \\ \midrule
\end{tabular}}
\vspace{-1.2\baselineskip}
\end{table}

\paragraph{Computing Resource} Collecting a group of pseudo datasets takes about 10 hours on a single 4090, while using 12GB GPU memory. The training times for single displacement and combination bond are approximately 6 hours and 1.5 hours, respectively, on a single 4090, and it only requires 3GB of GPU memory. Tuning the displacement product consumes 15 hours on single 4090.

\section{Conclusion}
This paper proposes \textbf{FreeBind} to enhance pre-trained unified multimodal representations by binding the knowledge of extra expert spaces. Based on the concept of viewing multimodal spaces as basic unit, we design two basic ``space bonds": displacement and combination bond. With these foundations, we introduce complex sequential \& parallel bonds to effectively combine multiple spaces simultaneously. After training, a coarse-to-fine customized inference strategy is employed to flexibly enhance unified space for different applications. Experimentally, we integrate ImageBind's audio-image-text space with multiple advanced spaces. The resulting space: ImageBind++, InternVL$_{I\!B}$ and InternVL$_{I\!B}$++ comprehensively surpass ImageBind. Moreover, via customized inference, it even outperforms state-of-the-art image-text and audio-text expert models in their respective domains.

\section{Impact Statements}
FreeBind enables flexible augment pre-trained unified space with very limited computing resources. Under appropriate usage, this technique can help quickly develop stronger unified multimodal representation with little training costs, and provide a powerful and accessible foundation for different customized multimodal application scenarios. However, low-cost unified representation learning methods could be misused to support unethical multi-modal applications. To prevent this, we plan to add unethical data detection to the pseudo dataset collection stage, thereby preventing representations from acquiring capabilities related to unethical applications.

\nocite{langley00}

\bibliography{example_paper}
\bibliographystyle{icml2024}

\newpage
\appendix
\onecolumn
\section{Pseudo Dataset between Unified and Audio-Text Expert Spaces}
Considering the source embeddings: $\mathbf{T}^{at}$, $\mathbf{A}^{at}$, $\mathbf{T}^u$, $\mathbf{V}^u$ and $\mathbf{A}^u$, the pseudo dataset starting from texts (i.e., $D_T$) can be expressed as:
\begin{equation}
\begin{gathered}
    \label{eq:over-data}
    \tilde{\mathbf{T}}^{at} = \mathbf{T}^{at}; \ \ \ 
    \tilde{\mathbf{A}}^{at} = \operatorname{softmax}(\tilde{\mathbf{T}}^{at} {\mathbf{A}^{at}}^\top)\mathbf{A}^{at};
    \\
    \tilde{\mathbf{T}}^u  = \mathbf{T}^u; \ \ \ 
    \tilde{\mathbf{A}}^{u} = \operatorname{softmax}(\tilde{\mathbf{T}}^{at} {\mathbf{A}^{at}}^\top) \mathbf{A}^{u}; \ \ \ 
    \tilde{\mathbf{V}}^{u} = \operatorname{softmax}(\tilde{\mathbf{T}}^u {\mathbf{V}^{u}}^\top) \mathbf{V}^{u}
\end{gathered}
\end{equation}
The pseudo dataset from audios (i.e., $D_A$) can be expressed as:
\begin{equation}
\begin{gathered}
    \label{eq:over-data}
    \tilde{\mathbf{A}}^{at} = \mathbf{A}^{at}; \ \ \ 
    \tilde{\mathbf{T}}^{at} = \operatorname{softmax}(\tilde{\mathbf{A}}^{at} {\mathbf{T}^{at}}^\top)\mathbf{T}^{at}; \\
    \tilde{\mathbf{A}}^u  = \mathbf{A}^u; \ \ \ 
    \tilde{\mathbf{T}}^{u} = \operatorname{softmax}(\tilde{\mathbf{A}}^{at} {\mathbf{T}^{at}}^\top) \mathbf{T}^{u}; \ \ \ 
    \tilde{\mathbf{V}}^{u} = \operatorname{softmax}(\tilde{\mathbf{A}}^u {\mathbf{V}^{u}}^\top) \mathbf{V}^{u}
\end{gathered}
\end{equation}

The pseudo dataset from non-shared image modality (i.e., $D_V$) can be expressed as:
\begin{equation}
\begin{gathered}
    \label{eq:over-data}
    \tilde{\mathbf{V}}^u = \mathbf{V}^u; \ \ \ 
    \tilde{\mathbf{T}}^{u}\! =\! \operatorname{softmax}(\tilde{\mathbf{V}}^u {\mathbf{T}^{u}}^\top\!)\mathbf{T}^{u}; \ \ \ 
    \tilde{\mathbf{A}}^{u}\! =\! \operatorname{softmax}(\tilde{\mathbf{V}}^u {\mathbf{A}^{u}}^\top\!) \mathbf{A}^{u}; \\
    \tilde{\mathbf{T}}^{at}\! =\! \operatorname{softmax}(\tilde{\mathbf{V}}^u {\mathbf{T}^{u}}^\top\!)\mathbf{T}^{at}; \ \ \ 
    \tilde{\mathbf{A}}^{at}\! =\! \operatorname{softmax}(\tilde{\mathbf{V}}^{u} {\mathbf{A}^{u}}^\top\!) \mathbf{A}^{at}
\end{gathered}
\end{equation}

\section{Training Datasets}

\paragraph{Unimodal data}
Following~\cite{wang2023connecting}, we employ the texts of COCO~\cite{lin2014microsoft}, CC3M~\cite{changpinyo2021conceptual, sharma2018conceptual}, MSRVTT~\cite{xu2016msr}, MAD~\cite{soldan2022mad}, AudioCaps~\cite{kim2019audiocaps} and Clotho~\cite{drossos2020clotho} as the unimodal source text. There are 2.33M text samples in total (only 1M texts are selected from CC3M). All the unpaired image data are from ImageNet~\cite{deng2009imagenet} training set, which consists of 1.3M images without any annotations. The audios are sourced from AudioSet~\cite{gemmeke2017audio} training set, total in 2M audio clips.

\paragraph{Paired data} 
Optionally, we utilize the 2 million audio-image pairs from the unbalanced training set of AudioSet to tune the audio encoder for the displacement bond product. Notably, AudioSet is also the training set of ImageBind. Therefore, utilizing AudioSet for tuning does not introduce any new knowledge. The purpose of further tuning is to repair the representation damage caused by the displacement bond process.

\section{Further Analysis of Combining Factors}
\label{sec:further}
\begin{figure*}[t]
	\centering
	\includegraphics[width=\linewidth]{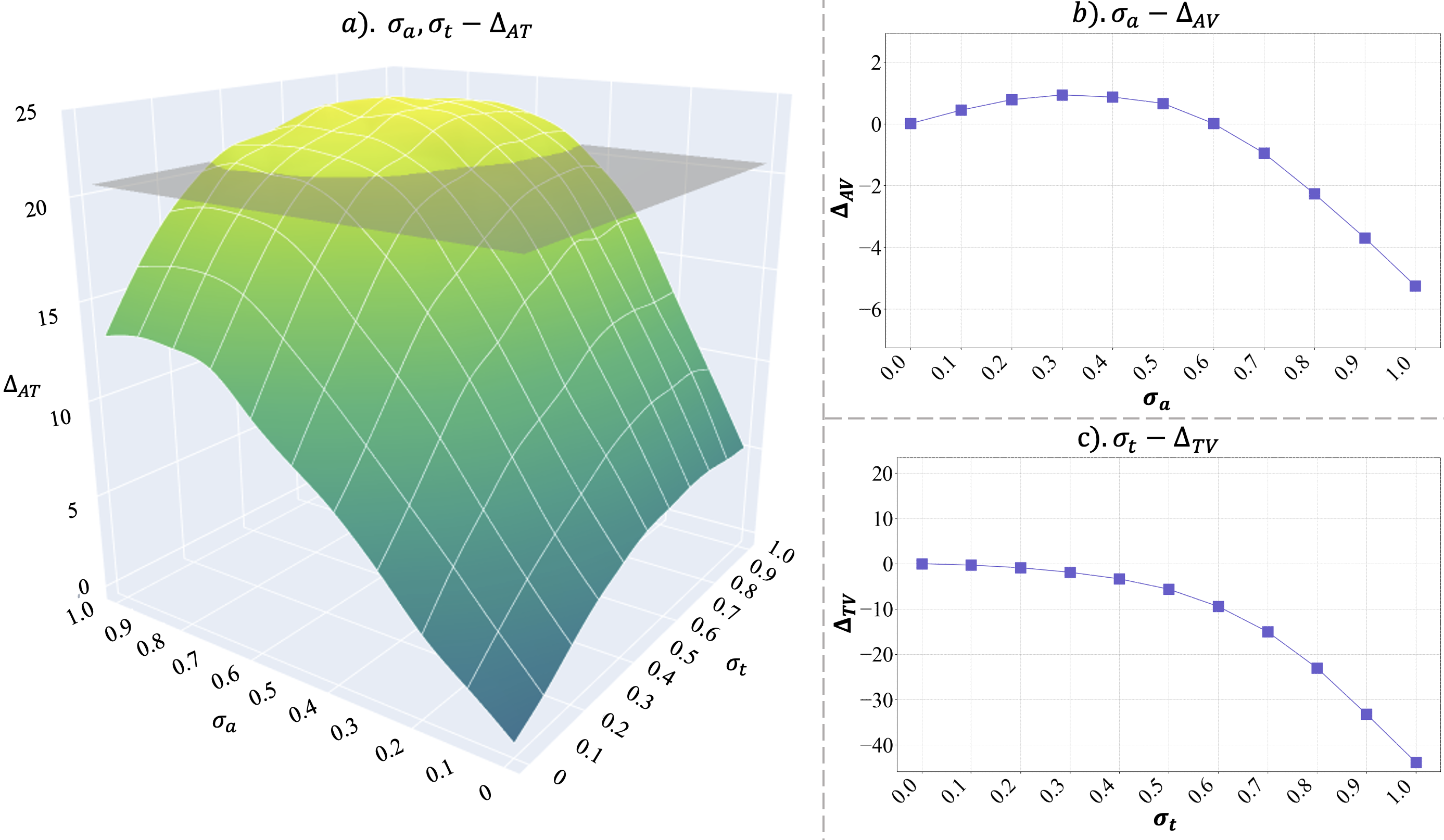}
        \vspace{-1\baselineskip}
	\caption{Analysis of CLAPs' combining factors ($\sigma_a, \sigma_t$) on InternVL$_{I\!B}$++. $\Delta_{AT}, \Delta_{A\!V}, \Delta_{TV}$ represents the average R@1 variance between InternVL$_{I\!B}$++ and InternVL$_{I\!B}$ on audio-text, audio-image and image-text retrieval tasks, respectively. The gray plane in the 3D figure $a)$ denotes the audio-text performance of CLAP$_{g}$. }
        \vspace{-1\baselineskip}
        \label{fig:analysis_VL++}
\end{figure*}

\begin{figure*}[t]
	\centering
	\includegraphics[width=\linewidth]{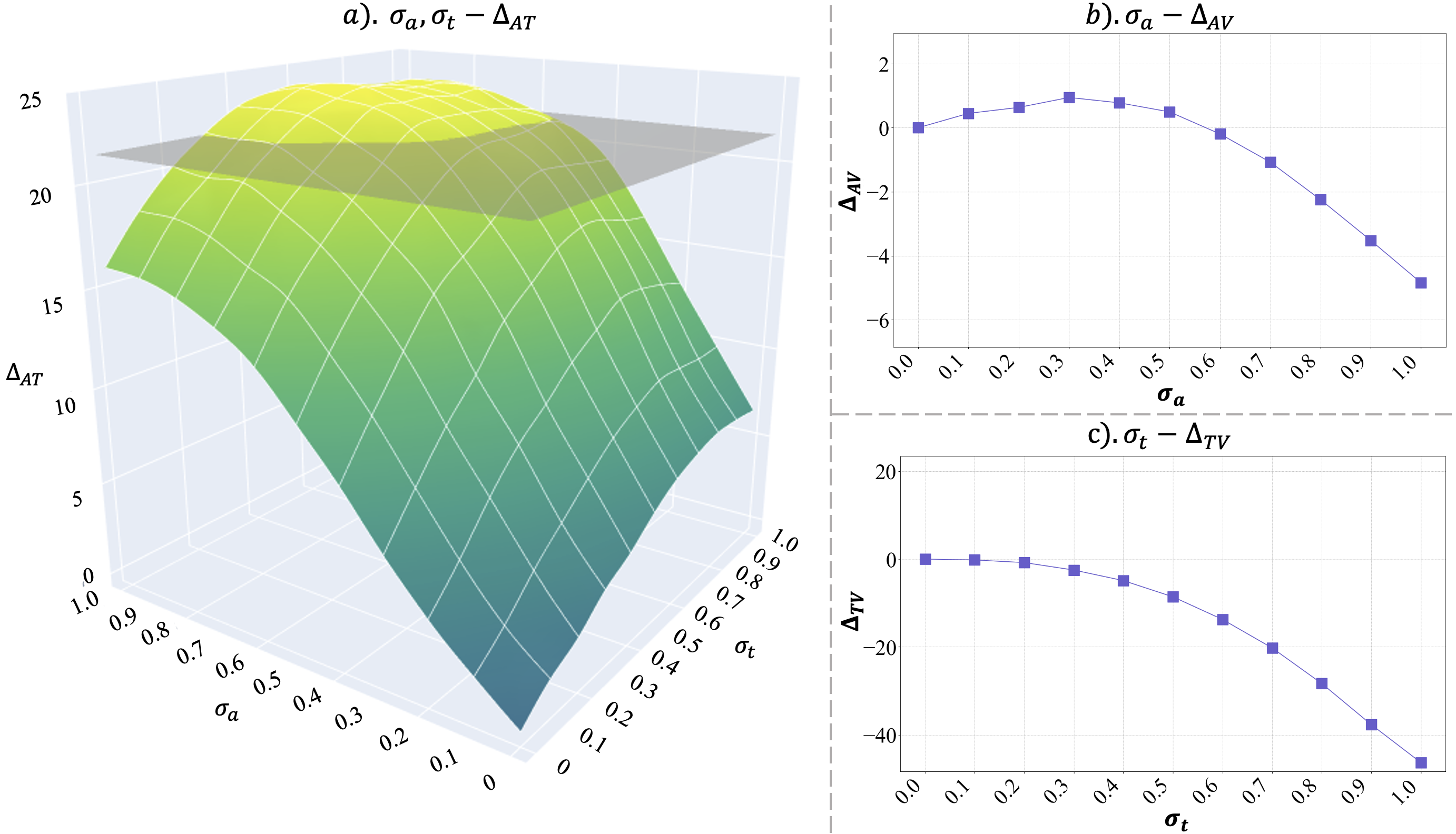}
        \vspace{-1\baselineskip}
	\caption{Analysis of CLAPs' combining factors ($\sigma_a, \sigma_t$) on ImageBind++. $\Delta_{AT}, \Delta_{A\!V}, \Delta_{TV}$ represents the average R@1 variance between ImageBind++ and ImageBind on audio-text, audio-image and image-text retrieval tasks, respectively. The gray plane in the 3D figure $a)$ denotes the audio-text performance of CLAP$_{g}$. }
        \vspace{-1\baselineskip}
        \label{fig:analysis_IB++}
\end{figure*}

To more comprehensively demonstrate the impact of the CLAP's combining factors on the product, we also analyzed CLAPs' combining factors ($\sigma_a, \sigma_t$) on InternVL$_{IB}$++ and ImageBind++, which are presented in Figure~\ref{fig:analysis_VL++} and~\ref{fig:analysis_IB++}. The curves and surfaces in these figures are similar to that of Figure~\ref{fig:analysis}. This observation further demonstrates the regularity and insensitivity of combining factors, as discussed in Section~\ref{sec:disc}.

\begin{figure*}[t]
	\centering
	\includegraphics[width=\linewidth]{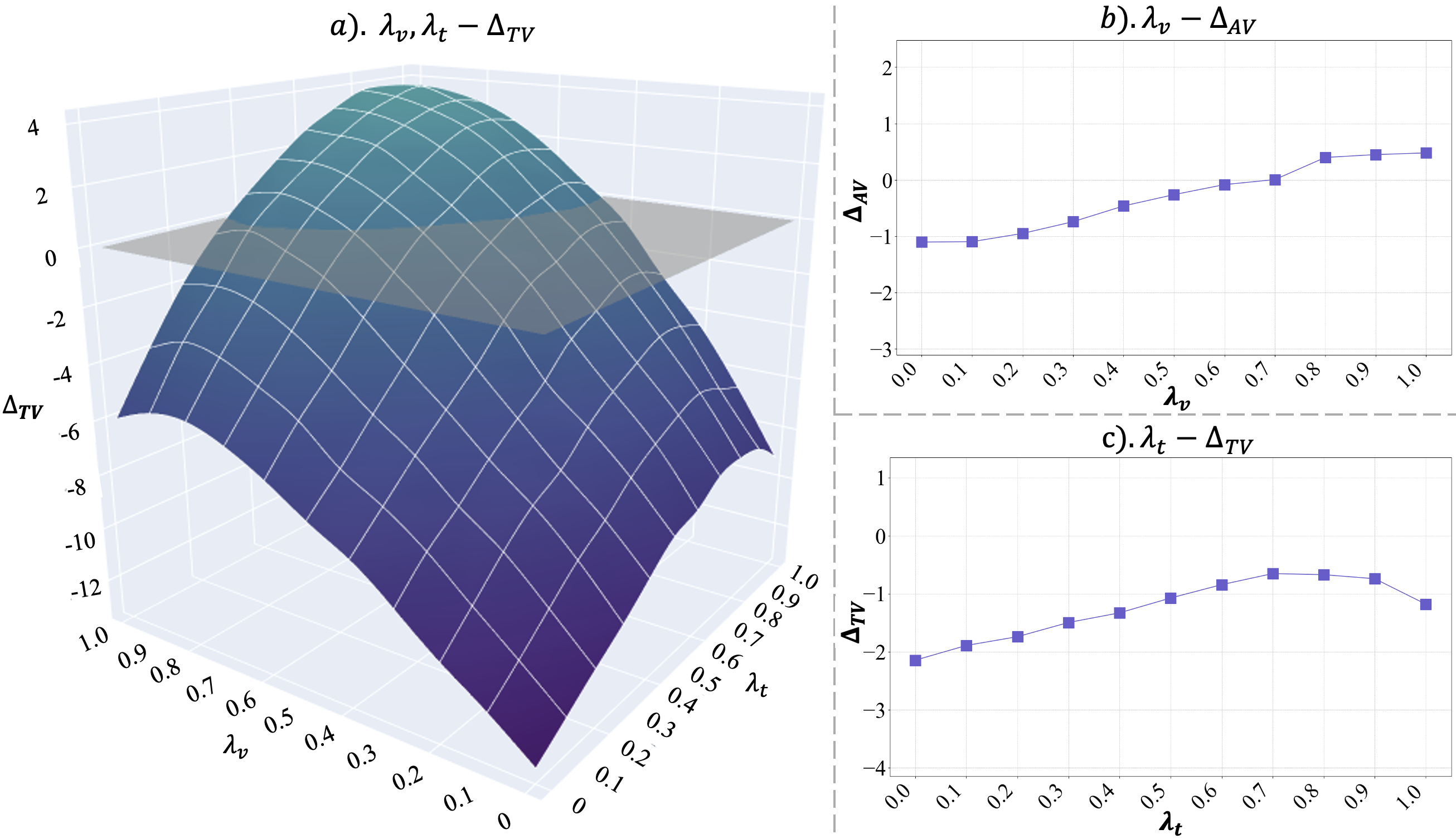}
        \vspace{-1\baselineskip}
	\caption{Analysis of InternVL's combining factors ($\lambda_v, \lambda_t$) on InternVL$_{I\!B}$. $\Delta_{AT}, \Delta_{A\!V}, \Delta_{TV}$ represents the average R@1 variance between InternVL$_{I\!B}$ and ImageBind on audio-text, audio-image and image-text retrieval tasks, respectively. Positive $\Delta_{*}$ signifies improvements in the corresponding task, while negative values indicate reductions. The gray plane in the 3D figure $a)$ denotes the image-text performance of ImageBind. }
        \vspace{-1\baselineskip}
        \label{fig:analysis_VL}
\end{figure*}

Moreover, we further display the impact of the InternVL's combining factor ($\lambda_t$, $\lambda_v$) on the performance of InternVL$_{IB}$ in Figure~\ref{fig:analysis_VL}. Generally speaking, since ImageBind's representations are remapped, the greater ($\lambda_t$, $\lambda_v$), the higher the overall performance, which is also consistent with the definition of displacement.


\section{Limitations and Future Work}
This paper introduces FreeBind, a promising and cost-effective unified space augmentation and knowledge fusion solution, and provides an in-depth and comprehensive analysis and discussion of the key design. However, the current FreeBind is only utilized to enhance the most basic unified audio-image-text space, whereas the most advanced unified space methods, such as ImageBind and LanguageBind, have achieved unified representations of six or seven modalities. Further research to incorporate FreeBind for more modalities would be an interesting direction.

In light of our experiments on displacement bond, which have demonstrated its capability to substitute a stronger image-text space for the unified space and effectively repair the lost knowledge through tuning, and combination bonds with small combining factors can yield an enhanced unified space with stable gains and no negative consequences. FreeBind shows promise for broader applications.



\end{document}